# On the Importance of Image Encoding in Automated Chest X-Ray Report Generation

Otabek Nazarov
otabek.nazarov@mbzuai.ac.ae

Mohammad Yaqub
mohammad.yaqub@mbzuai.ac.ae

Karthik Nandakumar
karthik.nandakumar@mbzuai.ac.ae

Mohammed Bin Zayed University of Artificial Intelligence
Abu Dhabi, UAE

## Abstract

Chest X-ray is one of the most popular medical imaging modalities due to its accessibility and effectiveness. However, there is a chronic shortage of well-trained radiologists who can interpret these images and diagnose the patient's condition. Therefore, automated radiology report generation can be a very helpful tool in clinical practice. A typical report generation workflow consists of two main steps: (i) encoding the image into a latent space and (ii) generating the text of the report based on the latent image embedding. Many existing report generation techniques use a standard convolutional neural network (CNN) architecture for image encoding followed by a Transformer-based decoder for medical text generation. In most cases, CNN and the decoder are trained jointly in an end-to-end fashion. In this work, we primarily focus on understanding the relative importance of encoder and decoder components. Towards this end, we analyze four different image encoding approaches: direct, fine-grained, CLIP-based, and Cluster-CLIP-based encodings in conjunction with three different decoders on the large-scale MIMIC-CXR dataset. Among these encoders, the cluster CLIP visual encoder is a novel approach that aims to generate more discriminative and explainable representations. CLIP-based encoders produce comparable results to traditional CNN-based encoders in terms of NLP metrics, while fine-grained encoding outperforms all other encoders both in terms of NLP and clinical accuracy metrics, thereby validating the importance of image encoder to effectively extract semantic information. GitHub repository: https://github.com/mudabek/encoding-cxr-report-gen

## 1 Introduction

Chest X-ray is a commonly used medical imaging modality because it covers a wide variety of diseases occurring in the chest area and the process of X-ray image acquisition is simple and efficient. However, there is a shortage of skilled radiologists who can interpret data in a timely manner and this issue became even more apparent during the COVID-19 outbreak. Automated report generation using machine learning can alleviate the problem of shortage of radiologists. The goal of automated radiology report generation is to produce an accurate report describing the patient's condition based on the given X-ray image. This task falls under the umbrella of image captioning algorithms. Though there are many existing image captioning algorithms, they are not directly applicable to the task of report generation. The main reason is that most of them produce short descriptions of various natural images, whereas radiology reports usually consist of several sentences describing fairly similar images with subtle but important differences.

In general, radiology report generation consists of two components: (i) an image encoder that produces an informative representation of the given image, and (ii) a medical text decoder that produces the report based on the information coming from the encoder. Typically, a Convolutional Neural Network (CNN)-based deep learning architecture [8] is used for image encoding. Earlier works mostly used a Long Short-Term Memory (LSTM) based recurrent neural network architecture [4, 5, 10, 11, 15] for decoding, whereas more recent methods are based on the Transformer architecture [2, 7, 17, 18, 23]. While it is obvious that image encoding plays a critical role in report generation, surprisingly it has received very little attention from the research community. Most recent works employ a standard CNN-based encoder and focus only on the decoder component. In this work, we show that image encoding plays a critical role in ensuring the accuracy of the generated reports. The contributions of this work are two fold:

- Compare different image encoding approaches (direct, fine-grained, CLIP, and Cluster-CLIP) along with multiple decoders to understand the relative importance of encoder and decoder components.

- Propose a novel cluster CLIP visual encoder (CCVE) that aims to generate more discriminative and explainable representations

## 2 Related Work

One of the first successful automated radiology report generation systems was based on the hierarchical LSTM model [10]. In this model, VGG-16 [20] was used as an image encoder and a hierarchical LSTM with an attention mechanism was used for decoding. The hierarchical LSTM approach was further refined in [4, 11] by creating two separate hierarchical LSTM models (one for normal and the other for abnormal cases). Modifying the image encoder was shown to improve the performance of the hierarchical LSTM model [15, 25]. While the CNN image encoder was pre-trained on the chest X-ray disease classification task [15], a special type of pooling operation was introduced in [25]. The common weakness of LSTM models is the lack of diversity in the generated reports due to possible reasons such as dataset imbalance and the tendency of LSTM models to overfit. Moreover, their efficacy was often evaluated based on natural language generation (NLG) metrics, which are not appropriate for the report generation task because even a random retrieval model can achieve good NLG metric results [1].

Another approach for report generation relies on utilizing existing reports in the database as templates. Li et. al. [12] created a hybrid model that recurrently chooses a sentence from a database or generates a new one using LSTM at the report generation stage. Syeda-Mahmood et. al. [22] used fine-grained template retrieval and achieved the highest NLG metrics reported thus far. Fine-grained report retrieval has good NLG metrics because it directly uses reports of radiologists. However, it does not have the ability to generalize well to cases not present in the database, which is the main weakness of the template-based approach.

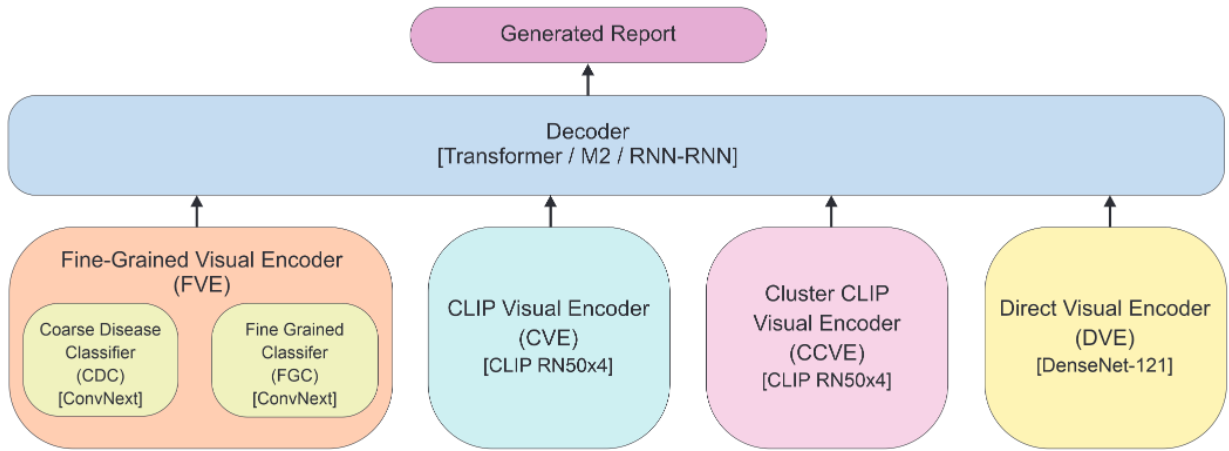

Figure 1: Illustration of various encoders and decoders used in this study.

Recently, Transformers [23] have gained prominence as the preferred architecture for the task of report generation. Chen et. al. [2] extracted image features using CNN and generated sentences using a Transformer decoder enhanced with relational memory. Hou et. al. [7] simply replaced the encoder of a vanilla Transformer with a DenseNet CNN encoder and generated reports with a plain Transformer decoder. Wang et. al. [24], Liu et. al. [14], and Najdenkoska et. al. [18] developed complex architectures incorporating BERT-based Transformers in various ways into their models. However, most of the existing methods [2, 4, 7, 10, 12, 13, 18] rely only on a simple CNN encoder trained jointly with a decoder for extracting information from the image. There is no existing work that analyzes image encoders in depth and quantifies their impact on the decoder. Hence, it becomes necessary to study the impact of different encoding strategies, which we have attempted in this work.

# 3   Proposed Method

Ideally, an image embedding must include all the semantic information that is necessary for generating the report. This requires training the image encoder and text decoder together in an end-to-end fashion. One of the issues with jointly training the image encoder and text decoder is that the cross-entropy loss function does not necessarily force the model to learn important semantic information for clinically accurate report generation. Rather the model often learns to generate the most frequent word sequences by focusing on common features present in most images discarding abnormality detection. With this in mind, we analyze techniques for more effective image feature extraction and evaluate their relative strengths.

The proposed system for automated radiology report generation is shown in Figure 1. It consists of image encoder and text decoder modules. Four types of image encoders are considered: a direct visual encoder (DVE), a fine-grained visual encoder (FVE), a CLIP visual encoder (CVE), and a cluster CLIP visual encoder (CCVE). FVE has a coarse disease classifier (CDC) for predicting 14 disease labels and a fine-grained classifier (FGC) for predicting 410 labels. CVE and CCVE are pre-trained to match the image embedding with the text embedding of reports. Each visual encoder produces an image embedding and these embeddings can either be fed individually or jointly to any type of decoder to produce the report in a recurrent manner.

## 3.1 Image Encoding

**Direct Visual Encoding (DVE)**: For direct encoding, we use the DenseNet-121 CNN architecture and train it end-to-end along with the decoder.

**Fine-Grained Visual Encoding (FVE)**: FVE consists of a coarse disease classifier (CDC) and a fine-grained classifier (FGC). Coarse disease labels can be automatically extracted from the radiology reports using the CheXpert library [9]. We obtained fine-grained labels by enhancing CheXpert with the capability to extract modifiers of the diseases (mild pneumonia, moderate cardiomegaly, etc.). We used the NLP spaCy library [6] to extract adjective and noun modifiers of a given disease. As a result, we obtained 410 fine-grained classes in total (each class occurs at least 100 times in the dataset). We use a ConvNeXt-small [16] CNN model as CDC and FGC. ConvNeXt-small is a recent state-of-the-art CNN architecture, which has shown better results in our experiments on the task of CXR image classification in comparison to other well-known CNN architectures such as DenseNet and ResNet. Feature embeddings from both CDC and FGC are used as inputs to the transformer decoder.

**CLIP Visual Encoding (CVE)**: CVE relies on contrastive language-image pretraining (CLIP) [19] to learn the image embedding for a given chest X-ray image. CLIP is a powerful multimodal model that has been trained on 400 million natural image-text pairs on the task of matching visual and textual embeddings of input data. This pre-trained model is available on OpenAI's GitHub page [19]. It consists of an image encoder module (various ResNet or Vision Transformer (ViT) models) and a text encoder (CLIP Transformer encoder) module.

We further fine-tune the pretrained CLIP model on our chest X-ray dataset by matching chest X-ray images with their corresponding reports. CLIP's text encoder can handle only 77 tokens and it uses binary pair encoding tokenization. However, radiology reports directly tokenized using binary pair encoding are usually longer than 77 tokens. Therefore, we could not use the radiology report directly as an input to CLIP's text encoder. Instead, we extract impression sections from the reports (if present) and use those as text input to CLIP. Impressions are usually one or two sentences long and contain the most important observation. Thus, the fine-tuned CLIP embedding can be expected to encode the most relevant information. After fine-tuning the CLIP model on the chest X-ray dataset, only the CLIP's visual encoder is used to obtain the embedding of the given image.

**Cluster CLIP Visual Encoding (CCVE)**: One of the limitations of CLIP is that the generated image embeddings for different images tend to be very similar, making it challenging for the decoder to differentiate well between them. To overcome this issue, we propose a novel cluster CLIP visual encoder (CCVE) with the goal of generating more discriminative image embeddings (see Figure 2). During the training phase, the CCVE module first clusters the reports in the database into 13 categories based on the impression section. Next, we create a set of 13 distinct convolution operators to act as filters. Given a CXR image and an impression with label $k$, we pass the image through the $k^{th}$ convolution filter and use CLIP to match the embedding of the filtered image with the text embedding of the impression. During the inference phase, the given image is processed using all the 13 filters and all the 13 filtered image embeddings are passed as inputs to the decoder.

The proposed CCVE method has two advantages. Firstly, it generates multiple diverse embeddings from the same image because the convolution filters are trained to focus on different aspects based on the disease category. Secondly, due to the transformer's self-attention mechanism, the attention output of the classification token of the transformer can reveal which of the 13 image embeddings of CCVE are most relevant for a given image. Filtered images corresponding to these embeddings can highlight areas important for report

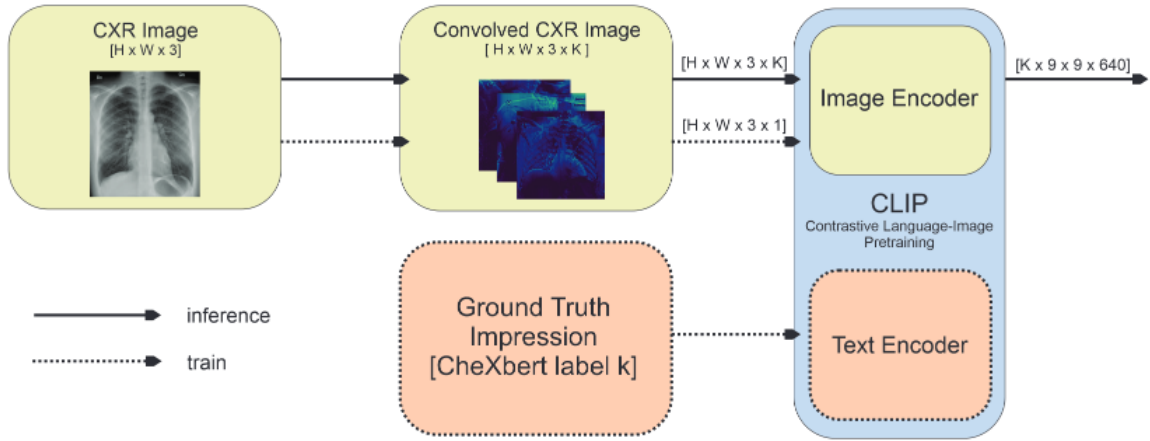

Figure 2: Cluster-based CLIP training and inference setup. $K = 13$ and corresponds to 13 disease labels from CheXbert labeler.

generation, thereby enhancing explainability.

The clustering of impression sections of the reports is performed using the CheXbert automatic report labeling library [21]. CheXbert uses named entity recognition NLP techniques along with BERT to automatically label sentences into one of the 14 CXR disease categories. We selected only 13 of them because two classes ('pleural effusion' and 'pleural other') were identifying almost the same sentences during parsing.

## 3.2 Report Text Generation

We use three different decoding methods for evaluating the effectiveness of image encoding techniques: vanilla transformer decoder [7], M2 (meshed-memory transformer) transformer decoder [17], and CNN-RNN-RNN decoder [15]. M2 transformer is a state-of-the-art captioning algorithm that has been further optimized by Miura et. al. [17] for the task of report generation. The CNN-RNN-RNN decoder relies on hierarchical RNN for decoding [15]. Each of these decoders receives outputs from FVE, CVE, CCVE, or DVE modules in the form of some latent representation of an X-ray image. It must be emphasized that there are differences in pre-processing steps, hyperparameter settings, and training strategies used in [7] and [17]. Since our primary goal is benchmarking various image encoders, we re-implemented the decoders ourselves to run the experiments under the same settings for a fair comparison. Hence, it is not possible to directly compare the results reported in this work with those reported in the literature [7, 17].

## 4 Experimental Results

### 4.1 Dataset

For our experiments, we used the MIMIC-CXR dataset, which is currently the largest publicly available radiology report dataset. It consists of 473,057 images and 206,563 reports from 63,478 patients. Among these images, there are 240,780 anteroposterior (AP), 101,379 posteroanterior (PA), and 116,023 lateral (LL) views. The associated reports consist of multiple sections: background information, impression, and findings. For our experiment, we have used only the AP images (other views are not consistently available for all patients) and retained only the findings sections of the reports as they contain the most information. We have used the default test set of AP images (3800 samples) provided within the MIMIC-CXR dataset in all our evaluations.

## 4.2 Implementation Details

All our experiments were carried out on an RTX A6000 48GB GPU using Python's PyTorch library.
**Fine-Grained Visual Encoding (FVE)**: We trained both CDC and FGC in the following way: batch size of 32, learning rate of 1e-4, early stopping with patience of 10, and ImageNet weight initialization. The model hyperparameters were picked based on the average AUC maximization on the held-out validation set. Robust deep AUC maximization (DAM) loss was used for network optimization. During the training process of decoders, the FVE was frozen and only used for the generation of coarse and fine-grained image embeddings with the dimension of $7 \times 7 \times 768$ each.
**CLIP Visual Encoding (CVE) and Cluster CLIP Visual Encoding (CCVE)**: Fine-tuning the CLIP model requires a carefully selected set of hyperparameters because it is prone to gradient explosion. We selected ResNet(RN)50x4 as the visual backbone and the following set of hyperparameters provided the best performance: learning rate of 1e-6, batch size of 64, and Adam optimizer with weight decay of 0.2. We trained the model for 30 epochs and selected the epoch with the lowest loss on the validation set. The same training set was used in the case of CCVE. As in the case of FVE, CVE and CCVE were frozen during the training process for decoders and their image embeddings with sizes of $9 \times 9 \times 640$ and $13 \times 640$, respectively, were passed to the decoders.
**Direct Visual Encoding (DVE)**: DenseNet-121 produces features of size $7 \times 7 \times 1024$, which are passed to the decoders for end-to-end training.
**Report Text Generation**: All of the decoders were trained with the same set of hyperparameters: 15 epochs, batch size of 24, and learning rate of 5e-4 with Adam optimizer.

## 4.3 Results and Discussion

The performance of all encoder-decoder configurations is reported in Table 1. Performance has been measured using both NLP metrics (BLEU, ROUGE, METEOR, and CIDER) and clinical accuracy metrics (precision, recall, and F1). It can be observed that simple CNN-based image encoders such as DVE and FVE generally outperform the contrastively trained CLIP-based encoders (CVE and CCVE). This is especially true in the case of clinical accuracy metrics. Furthermore, FVE along with transformer decoders has demonstrated the best results across various NLP and clinical accuracy metrics. Some examples of the generated reports are shown in Figure 3. Regardless of the decoder, DVE, CVE, and CCVE encoders produce similar reports. However, FVE captures most of the abnormalities present in the image, which is the most important aspect of report generation.

We hypothesize that the relatively poor performance of CLIP-based encoders (CVE and CCVE) can be attributed to the limitations in the chosen training strategy. Recall that DVE and FVE are trained using the text in the findings section of the report. While DVE is trained end-to-end, FVE is trained based on coarse and fine-grained labels extracted from the findings section. On the other hand, due to issues with tokenization length, CLIP fine-tuning is based on the impressions section of the report. Though the impression section is supposed to summarize the most relevant information, the impressions section of the reports have numerous common words and very few words identifying the abnormality. Consequently, we believe that the CLIP-based encoders tend to focus on the common words and features between the samples. Moreover, the use of CheXbert labeler in CCVE is likely to introduce errors in the clustering process, which can propagate through the subsequent models during training. This highlights the need for a better training strategy to fine-tune the CLIP model.

Table 1: Performance of automated X-ray report generation models. Here, B, RG, MTR, CDR, P, and R correspond to BLEU, ROUGE, METEOR, CIDER, precision, and recall metrics, respectively and F1 is the mean of all the per-class F1 scores.

| Model | B1 | B2 | B3 | B4 | RG | MTR | CDR | P | R | F1 |
|---|---|---|---|---|---|---|---|---|---|---|
| Transformer Decoder | | | | | | | | | | |
| DVE | 0.286 | 0.172 | 0.115 | 0.083 | 0.231 | 0.116 | 0.109 | 0.320 | 0.179 | 0.169 |
| CCVE | 0.267 | 0.159 | 0.104 | 0.074 | 0.224 | 0.107 | 0.091 | 0.246 | 0.142 | 0.108 |
| CVE | 0.276 | 0.165 | 0.110 | 0.079 | 0.221 | 0.110 | 0.092 | 0.382 | 0.142 | 0.129 |
| FVE | **0.299** | **0.182** | **0.124** | **0.090** | **0.238** | **0.123** | **0.136** | **0.443** | **0.212** | **0.220** |
| M2 Decoder | | | | | | | | | | |
| DVE | 0.297 | 0.181 | 0.123 | 0.089 | 0.238 | 0.123 | 0.129 | **0.418** | 0.205 | 0.211 |
| CCVE | 0.266 | 0.159 | 0.105 | 0.073 | 0.224 | 0.108 | 0.090 | 0.249 | 0.146 | 0.130 |
| CVE | 0.278 | 0.167 | 0.112 | 0.081 | 0.227 | 0.112 | 0.103 | 0.206 | 0.345 | 0.116 |
| FVE | **0.298** | **0.183** | **0.124** | **0.090** | **0.242** | **0.125** | **0.137** | 0.402 | **0.232** | **0.236** |
| RNN-RNN Decoder | | | | | | | | | | |
| DVE | **0.289** | **0.171** | **0.114** | **0.081** | 0.228 | 0.112 | 0.112 | 0.296 | 0.163 | 0.153 |
| CCVE | 0.246 | 0.147 | 0.097 | 0.068 | 0.225 | 0.104 | 0.096 | 0.243 | 0.137 | 0.103 |
| CVE | 0.254 | 0.152 | 0.101 | 0.071 | 0.226 | 0.106 | 0.096 | **0.344** | 0.140 | 0.100 |
| FVE | 0.277 | 0.167 | 0.112 | 0.080 | **0.235** | **0.116** | **0.116** | 0.309 | **0.187** | **0.172** |

### 4.3.1 CCVE Evaluation

To further evaluate CCVE, we ran both qualitative and quantitative tests. For quantitative analysis, we extracted image embeddings, applied dimensionality reduction using PCA (from 8320 to 2000 keeping approximately 70% of the variance), and fed them to the CatBoost classifier [3]. The model gave the ROC-AUC of 0.711 on the task of disease classification, which is a lower score when compared to CNN models from Table 2. Lower ROC-AUC based on CCVE features when compared to CNN models explains the lower macro-F1 score of CCVE on the task of report generation. The qualitative evaluation of CCVE using the t-SNE dimensionality reduction algorithm in Figure 4 shows the effectiveness of CCVE in producing distinct clusters for images, which was the initial motivation for creating this encoder. Finally, explainability is an important aspect of medical imaging and the better explainability of the proposed CCVE method is demonstrated in Figure 4, where the filtered images highlight areas in the image that are important for report generation.

### 4.3.2 Choice of FVE backbone

ConvNeXt-small was experimentally selected as the FVE backbone based on an architectural search. DenseNet or ResNet has been the default image encoder in many report generation architectures. However, we evaluated other options for the FVE backbone: ResNet-(18/50), DenseNet-121, ViT-small, Swin-small, ConvNeXt-small/base. All these encoders were initialized with ImageNet weights. Note that Swin-small and ViT-small were trained with different hyperparameters (batch size 256, learning rate 1e-4) and optimizer (SAM optimizer produced better results than Adam in our experiments) compared to the hyperparameters reported for CNN in Section 4.2. Even though ConvNeXt-small has the smallest number of parameters among all the above encoders, it had the best performance when evaluated on the task of image classification (see Table 2). CDC with a ConvNeXt-small backbone had ROC-AUC of 0.829 and FGC with the same backbone had ROC-AUC of 0.816 on the held-out validation set.

| Decoder | DVE | FVE | CVE | ICVE |
|---|---|---|---|---|
| Ground truth | **Monitoring and support devices are constant.** Constant cardiomegaly with relatively extensive retrocardiac atelectasis and the potential presence of a **small left pleural effusion.** Mild pulmonary edema . Areas of atelectasis at the right lung base. No newly occurred parenchymal opacities. **No pneumothorax.** | **Monitoring and support devices are constant.** Constant cardiomegaly with relatively extensive retrocardiac atelectasis and the potential presence of a **small left pleural effusion.** Mild pulmonary edema . Areas of **atelectasis** at the right lung base. No newly occurred parenchymal opacities. **No pneumothorax.** | **Monitoring and support devices are constant.** Constant cardiomegaly with relatively extensive retrocardiac atelectasis and the potential presence of a small left pleural effusion. Mild pulmonary edema . Areas of atelectasis at the right lung base. **No newly occurred parenchymal opacities.** No pneumothorax. | Monitoring and support devices are constant. Constant cardiomegaly with relatively extensive retrocardiac atelectasis and the potential presence of a small left pleural effusion. Mild pulmonary edema . Areas of atelectasis at the right lung base. No newly occurred parenchymal opacities. **No pneumothorax.** |
| Transformer | The patient has received a nasogastric tube. The course of the tube is unremarkable, the tip of the tube projects over the middle parts of the stomach. There is no evidence of complications, notably **no pneumothorax. No other monitoring and support devices are in unchanged position.** | **The monitoring and support devices remain in place. Continued enlargement of the cardiac silhouette** with some elevation of pulmonary venous pressure . **Opacification** at the left base is consistent with **volume loss** in the **left lower lobe and pleural effusion.** | The patient has been intubated. The tip of the endotracheal tube projects _ cm above the carina. The tube should be pulled back by approximately _ cm. The course of the nasogastric tube is unremarkable. There is no evidence of complications, notably **no pneumothorax.** | The patient has been intubated. The tip of the endotracheal tube projects _ cm above the carina. The course of the nasogastric tube is unremarkable, the tip of the tube projects over the middle parts of the stomach. There is no evidence of complications, notably **no pneumothorax.** Otherwise, the radiograph is |
| M2 | The patient has been intubated. The tip of the endotracheal tube projects cm above the carina. The course of the nasogastric tube is unremarkable, the tip of the tube projects over the middle parts of the stomach. There is no evidence of complications, notably **no pneumothorax.** Otherwise , the radiograph is unchanged . | **The monitoring and support devices remain in place. Continued enlargement of the cardiac silhouette** with evidence of elevated pulmonary venous pressure. Opacification at the right base is consistent with **volume loss** in the **left lower lobe and pleural effusion.** | The patient has received a nasogastric tube. The course of the tube is unremarkable, the tip of the tube projects over the middle parts of the stomach. There is no evidence of complications, notably **no pneumothorax. The other monitoring and support devices are unchanged.** | The patient has received a nasogastric tube. The course of the tube is unremarkable, the tip of the tube projects over the middle parts of the stomach. There is no evidence of complications, notably **no pneumothorax.** Otherwise , the radiograph is unchanged . |
| RNN-RNN | The patient has been intubated. The course of the nasogastric tube is unremarkable, the tip of the tube projects over the middle parts of the stomach. There is no evidence of complications, notably **no pneumothorax. The other monitoring and support devices are constant .** | The patient has been **intubated.** The course of the nasogastric tube is unremarkable , the tip of the tube projects over the middle parts of the stomach . There is no evidence of complications , notably **no pneumothorax.** Otherwise , the radiograph is unchanged. | There is no relevant change. The monitoring and support devices are in constant position. Unchanged appearance of the cardiac silhouette. **Unchanged appearance of the lung parenchyma.** No evidence of pneumonia. | The patient has received a nasogastric tube. The course of the tube is unremarkable, the tip of the tube projects over the middle parts of the stomach. There is no evidence of complications, notably **no pneumothorax.** Otherwise, the radiograph is unchanged. |

Figure 3: Examples of reports generated by different model configurations. The bold text highlights abnormalities that models detect. FVE is able to identify most of the abnormalities.

Table 2: ROC-AUC of models on the task of CXR image classification of 14 disease labels.

| Models | RN-18 | RN-50 | DN-121 | ViT-s | Swin-s | ConvNeXt-s | ConvNeXt-b |
|---|---|---|---|---|---|---|---|
| ROC-AUC | 0.808 | 0.810 | 0.817 | 0.720 | 0.762 | 0.829 | 0.810 |

### 4.3.3 Importance of Fine-Grained Labels

FVE is trained to encode the semantic information from the images. To further validate the importance of providing semantic information to the decoder, we used ground truth fine-grained labels as inputs to the encoder-decoder transformer. We fed ground truth fine-grained labels in the text form to the encoder. This configuration had a macro-F1 score of 0.778 on its own (see Table 3), demonstrating that semantic information provided by fine-grained labels is key to accurate report generation. This result indicates that semantic information extraction must be a priority for clinically accurate report generation.

Table 3: Performance of language encoder-decoder transformer with ground-truth fine-grained labels used as textual input.

| B1 | B2 | B3 | B4 | RG | MTR | CDR | P | R | F1 |
|---|---|---|---|---|---|---|---|---|---|
| 0.318 | 0.263 | 0.222 | 0.196 | 0.287 | 0.161 | 0.219 | 0.827 | 0.747 | 0.778 |

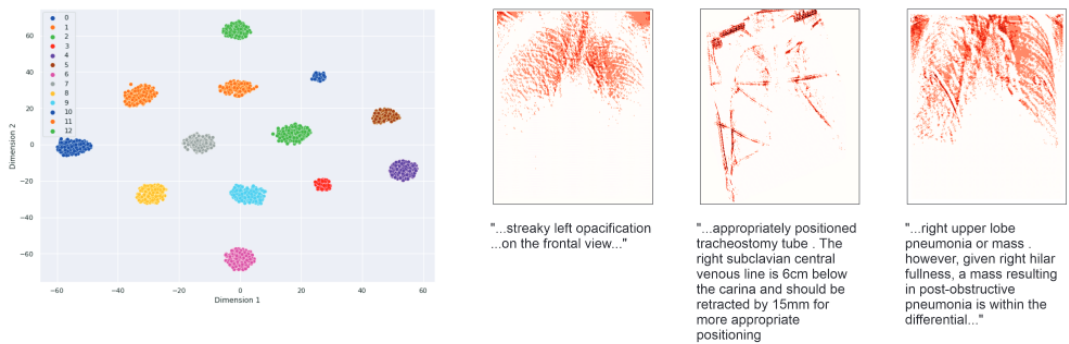

Figure 4: Left: t-SNE visualization of 13 distinct clusters of CCVE image embeddings. Right: Examples of outputs of CCVE's convolution operation - filtered images highlight relevant areas related to the disease mentioned in the report.

We have also experimented with using outputs of FVE in text form. We passed the fine-grained labels generated by FVE as input to the language encoder-decoder transformer. However, the results were much worse than those reported in Table 3 because FVE does not always produce correct labels and the encoder-decoder network did not train well under noisy conditions. Therefore, we conclude that it is better to utilize the image embeddings of FVE rather than the labels generated by it.

# 5  Conclusion

Most existing automated X-ray report generation systems focus on text decoding techniques and often overlook image encoding. In this work, we have analyzed four different image encoding techniques in depth. Our experiments show that encoders which are good at semantic information extraction, are also good at producing reports with the best NLP and clinical accuracy metrics. Thus, effective report generation models must have an image encoder optimized for semantic information extraction.

Furthermore, we have proposed an explainable cluster-based CLIP visual encoder to improve the efficiency of the CLIP visual encoder at capturing useful information. Though, it has not shown improvement when compared to the original CLIP encoder, we believe that the limitation lies with the training process. In general, our work highlights the need for more research effort on the image encoding component as it can boost the accuracy of the generated reports and push automated radiology report generation systems closer to industry deployment.